\ifdefined\XeTeXversion
  
\fi
\documentclass[letterpaper,10pt,conference]{ieeeconf}
\usepackage{amsmath,amssymb}
\usepackage{graphicx}
\usepackage{booktabs}
\usepackage{multirow}
\usepackage{tabularx}
\usepackage{array}
\usepackage[table]{xcolor}
\usepackage{cite}
\usepackage{float}
\usepackage{stfloats}
\usepackage{url}
\usepackage{flushend}
\graphicspath{{figures/}{./}}
\definecolor{vfVehFill}{HTML}{E5F1FA}
\definecolor{vfRsuFill}{HTML}{FFF0DC}
\definecolor{vfPurpleFill}{HTML}{EEE8F7}
\definecolor{vfGreenFill}{HTML}{E5F4EA}
\definecolor{vfGrayFill}{HTML}{F3F4F6}

\newcommand{\method}{VeriFuse}

\newcommand{\vect}[1]{\boldsymbol{#1}}
\newcommand{\R}{\mathbb{R}}
\newcolumntype{Y}{>{\centering\arraybackslash}X}
\newcolumntype{L}{>{\raggedright\arraybackslash}X}

\title{\LARGE\bf
\method: Bounded Vision--Language Arbitration and Reason-Guided Refinement for Cooperative 3D Perception}

\author{{\normalsize Hongyi Lin$^{1,2}$, Yiyao Liu$^{3}$, Qi Kang$^{4}$, Heye Huang$^{5}$, Yang Liu$^{1}$, Haris Koutsopoulos$^{4}$, and Jinhua Zhao$^{2}$}\\[0.2ex]
{\small $^{1}$Tsinghua University \quad $^{2}$MIT \quad $^{3}$UESTC \quad $^{4}$Northeastern University \quad $^{5}$KAIST}%
}

\IEEEaftertitletext{\vspace{-6pt}}
\let\vfOriginalBibliography\thebibliography
\renewcommand{\thebibliography}[1]{%
  \vfOriginalBibliography{#1}%
  \setlength{\baselineskip}{8.5pt}%
  \setlength{\itemsep}{0pt}%
}

\begin{document}
\maketitle
\thispagestyle{empty}
\pagestyle{empty}

\begin{abstract}
Vision--language models (VLMs) have demonstrated strong scene understanding and semantic judgment across diverse tasks, but their appropriate role in cooperative perception remains unclear. Directly asking a VLM to regress 3D detections is unreliable and computationally expensive, whereas using it to select the output of a single source discards useful information from other agents. We introduce \method, a bounded arbitration framework for vehicle--infrastructure cooperative 3D detection. Each agent first produces detections independently. Around each vehicle and roadside proposal, \method{} generates source-conditioned geometric candidates and combines the original detections, their perturbations, and cross-source hypotheses into a unified candidate pool. A frozen VLM then chooses among three admissible actions: \textsc{Select} an adequate candidate; \textsc{Refine} an existing anchor when an object is supported but all candidates are geometrically inadequate; or \textsc{Reject} an unsupported infrastructure-only proposal. Experiments on the DAIR-V2X dataset show that \method{} achieves 0.494/0.357 cooperative 3D AP$_{50}$/AP$_{70}$ and limits the relative vehicle-side BEV AP$_{50}$ drop under a 300~ms delay to 1.7\%. Overall, \method{} assigns the VLM a clear and constrained role in cooperative perception: semantic reasoning resolves ambiguity among cross-agent hypotheses, while deterministic constraints determine the final 3D geometry.
\end{abstract}

\section{Introduction}
\label{sec:intro}
Autonomous vehicles inevitably encounter incomplete or unreliable observations caused by occlusion, adverse conditions, or sensor degradation. For example, an ego sensor cannot detect an object that is fully hidden by a truck or building. Vehicle--infrastructure cooperative perception alleviates these limitations by sharing observations from roadside units (RSUs), whose elevated viewpoints complement those of the vehicle~\cite{dairv2x}. Existing systems commonly exchange raw measurements or learned features~\cite{v2vnet,where2comm}. Although these approaches can recover objects that are invisible to an individual agent, they require compatible perception and communication models. Sharing detection results is easier to deploy, however, conventional late fusion cannot reliably handle conflicting boxes or roadside false positives.

VLMs can interpret scene context and judge whether a detection is visually plausible. This makes them attractive for resolving conflicts between agents. However, VLMs are not strong metric detectors. Existing evaluations show that they remain less precise in object localization and are too slow for real-time detection~\cite{zhou2026sotif}. Hybrid robotic systems therefore use VLMs for semantic interpretation and specialized models for spatial localization~\cite{paetzold2025vlmgist}. Following this principle, we use the VLM to reason over detector-generated candidates rather than replace the detectors~\cite{lin2026proxyselect}.

\begin{figure}[t]
\centering
\includegraphics[width=\columnwidth]{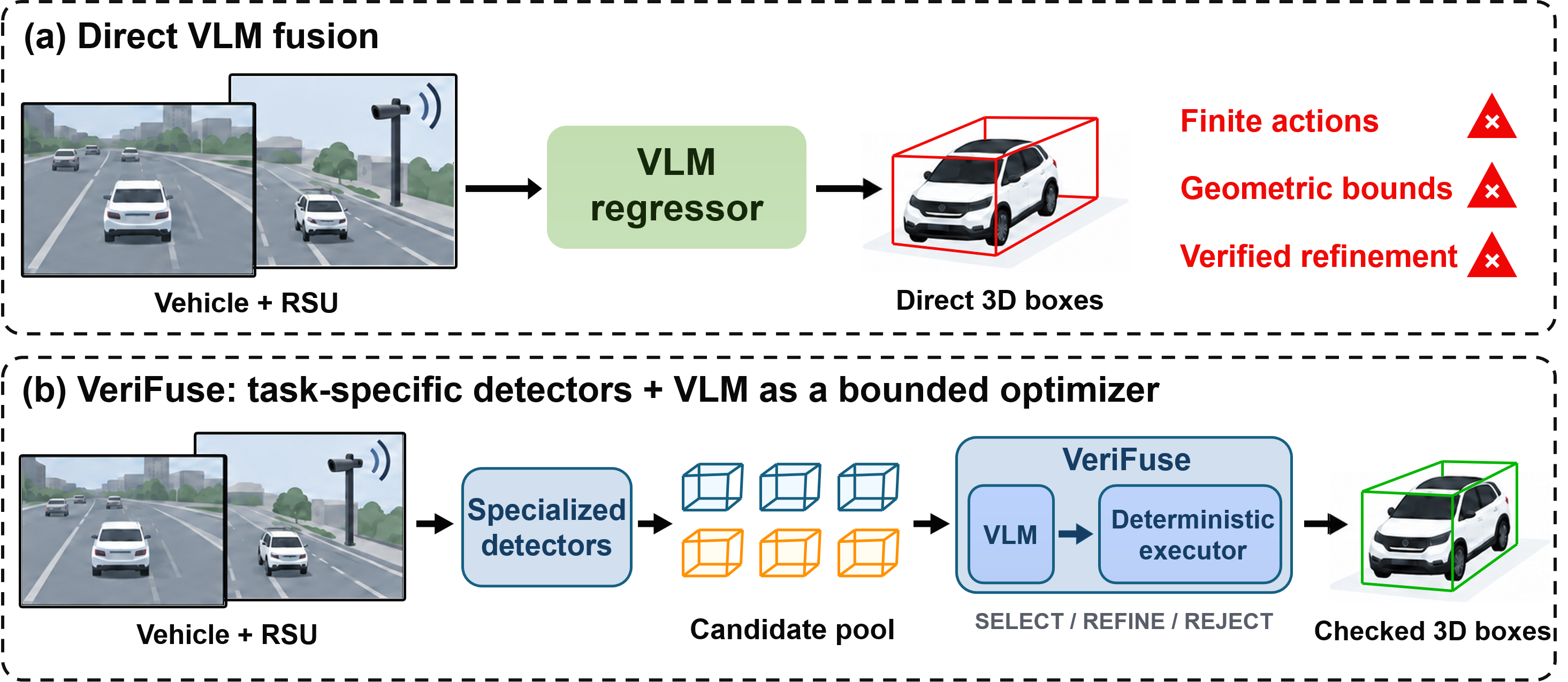}
\caption{Role of the VLM in cooperative 3D perception. (a)~Direct VLM regression predicts box coordinates directly. (b)~\method{} uses specialized detectors to generate metric hypotheses, a frozen VLM to select an action, and a deterministic executor to constrain the final box.}
\label{fig:comparison}
\end{figure}

Applying this principle to cooperative perception raises a more specific question: \emph{what should the VLM be allowed to decide?} Directly asking it to generate a 7-DoF box turns semantic reasoning into unconstrained metric regression and introduces substantial inference latency. Using it only to choose between the vehicle and RSU outputs is also insufficient, because a source-level decision discards potentially useful hypotheses from the other agent. Similarly, asking the VLM only whether an RSU detection should be retained reduces cooperative perception to binary validation. It cannot distinguish between an unsupported object and a real object for which both agents provide inaccurate boxes. Therefore, the role of the VLM must be defined at the hypothesis level rather than at the source level. Figure~\ref{fig:comparison} contrasts direct VLM regression with the proposed bounded arbitrator role.

We address these problems with \method, a bounded arbitration framework for cooperative 3D detection. The vehicle and RSU first generate detections independently. Since these detections are usually close to a plausible solution, \method{} constructs bounded candidates around each proposal using source-specific perturbation priors. The original boxes are retained, and compatible cross-source candidates are added to the same pool. A frozen VLM then chooses among three actions. \textsc{Select} accepts an adequate candidate, \textsc{Refine} corrects an inadequate candidate, and \textsc{Reject} removes an unsupported infrastructure-only proposal. For refinement, the VLM selects an anchor and identifies its error through a predefined reason code. A deterministic compiler then applies the corresponding bounded correction. This design uses VLM reasoning to resolve cross-agent conflicts while keeping 3D geometry under deterministic control. 

The contributions of this work are threefold:
\begin{itemize}
\item We establish a hybrid cooperative-perception architecture in which specialized detectors generate 3D proposals and a frozen VLM acts as a bounded optimizer.

\item We construct a unified source-conditioned candidate pool and define explicit \textsc{Select}, \textsc{Refine}, and \textsc{Reject} actions for different arbitration outcomes.

\item We introduce a reason-guided self-refining mechanism that corrects inadequate candidates without direct VLM coordinate regression.
\end{itemize}

\section{Related Work}
\label{sec:related}
\subsection{Cooperative 3D Perception}
Cooperative perception methods differ mainly in what agents exchange. Early fusion shares raw measurements, which preserves sensor evidence but requires high bandwidth and compatible sensor configurations~\cite{lin2025big}. Intermediate fusion shares learned features to support cross-agent aggregation~\cite{disconet}. Subsequent methods introduce Transformer-based fusion~\cite{v2xvit,cobevt} or selective communication~\cite{where2comm}. Other systems improve robustness to pose errors~\cite{coalign} and heterogeneous models~\cite{heal,stamp}. Although these methods can recover objects missed by a single detector, they depend on a shared or adapted feature protocol. Late fusion avoids this dependency by sharing detection boxes and applying NMS or box averaging~\cite{softnms,wbf}. Its interface is detector-independent, but fixed geometric rules cannot determine why two agents disagree. \method{} retains detection-list interoperability, requests bounded ROI evidence only when arbitration is needed, and uses that evidence to resolve conflicts among agent hypotheses.

\subsection{VLMs in Autonomous Driving}
VLMs have been applied to autonomous driving for scene interpretation and high-level reasoning. DriveLM formulates perception and planning as graph-structured visual question answering~\cite{drivelm}. DriveVLM uses a VLM to interpret complex scenes and generate hierarchical plans~\cite{drivevlm}. Its hybrid variant retains a conventional driving pipeline because direct VLM deployment is limited by spatial reasoning and computational cost. OmniDrive further develops 3D-aware driving reasoning through counterfactual supervision~\cite{omnidrive}. A SOTIF-oriented evaluation finds that VLMs provide useful semantic recall, whereas specialized detectors retain advantages in geometric precision and runtime under the tested conditions~\cite{zhou2026sotif}. These findings support combining VLMs with specialized perception models. Existing work, however, mainly considers single-agent understanding or planning. We instead study how a frozen VLM should arbitrate between conflicting outputs in cooperative perception.

\subsection{VLMs as Constrained Decision Makers}
An alternative to direct prediction is to constrain a general-purpose model to decisions that a specialized model can execute. VLM-GIST combines VLM-based semantic understanding with dedicated localization and tracking models~\cite{paetzold2025vlmgist}, while proxy-task reasoning asks a foundation model to select or verify hypotheses from a specialized predictor~\cite{lin2026proxyselect}. \method{} brings this separation of semantic judgment and metric execution to cooperative 3D perception: it arbitrates a mixed vehicle--RSU candidate pool, and a deterministic operator applies one bounded token-to-box update when no candidate is adequate.

\section{Problem Formulation}
\label{sec:problem}
At time $t$, each agent $a\in\mathcal A$ independently produces
\begin{equation}
\mathcal D_a^t=
\left\{d_{a,i}^t=(b_{a,i}^t,s_{a,i}^t,c_{a,i}^t)\right\}_{i=1}^{N_a},
\qquad b_{a,i}^t\in\R^7,
\label{eq:detections}
\end{equation}
where a box is parameterized as $b=(\vect p,\vect d,\psi)$, with center $\vect p=(x,y,z)$, positive dimensions $\vect d=(\ell,w,h)$, and yaw $\psi$. Calibration and timestamp alignment map every box to a common fusion frame, yielding $\bar b_{a,i}^t$.

Let $\nu=(a,i)$ identify an original proposal. At the fixed fusion time, we write $b_\nu\equiv\bar b_{a,i}^t$, $s_\nu\equiv s_{a,i}^t$, and $c_\nu\equiv c_{a,i}^t$, suppressing $t$ on proposal attributes while retaining it on temporal evidence and track states. Let $g$ denote an associated object cluster. Each cluster is source-unique, containing at most one proposal per source, and is assumed to represent at most one physical object. The association procedure is specified in Sec.~\ref{sec:association}. Timestamp-consistent sensor evidence available for the cluster is denoted by $E_g^t$.

For every original proposal $\nu$, let $\mathcal N_\nu(b_\nu)$ be its admissible geometric neighborhood. Because $E_g^t$ is fixed for each cluster, its dependence is suppressed in this notation. The feasible output set for cluster $g$ is
\begin{equation}
\begin{aligned}
\mathcal Y_g & =\{\varnothing\}\cup
\left\{(\widetilde b,\widetilde s,\widetilde c):
\widetilde b\in\mathcal N(g)\right\},\\
\mathcal N(g)&=\bigcup_{\nu\in g}\mathcal N_\nu(b_\nu).
\end{aligned}
\label{eq:feasible_output}
\end{equation}
The cooperative fusion problem is to construct $F(g,E_g^t)\in\mathcal Y_g$ under the \textsc{Select}, \textsc{Refine}, and \textsc{Reject} contract defined in Sec.~\ref{sec:arbitration}.

Equation~\eqref{eq:feasible_output} prevents unconstrained VLM box generation: every nonempty output remains in the neighborhood of at least one original detector proposal. The method selects or locally corrects detector-supported geometry; an object absent from every detection list lies outside its feasible output set. Under the main policy, a cluster containing a vehicle proposal cannot produce an empty output, whereas an infrastructure-only cluster may be suppressed when its evidence is unsupported.

\section{Methodology}
\label{sec:method}
\begin{figure*}[!t]
\centering
  \IfFileExists{figures/verifuse_overview_v2.pdf}{%
  \includegraphics[width=\textwidth]{figures/verifuse_overview_v2.pdf}%
}{%
  \IfFileExists{figures/verifuse_overview_v2.png}{%
    \includegraphics[width=\textwidth]{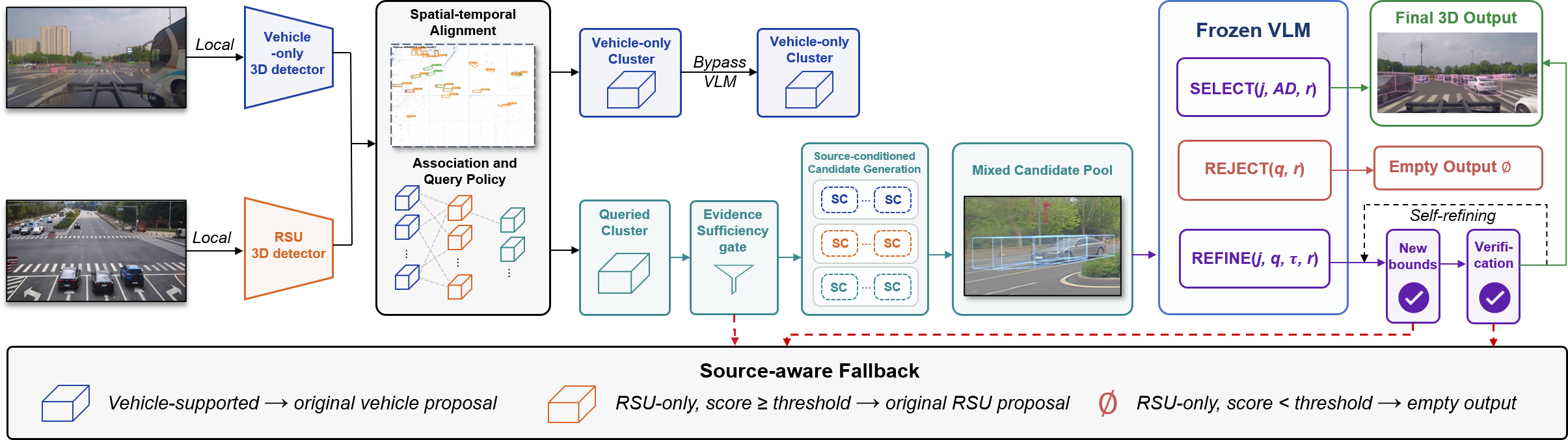}%
  }{%
    \fbox{\parbox[c][0.19\textheight][c]{0.96\textwidth}{\centering
    \textbf{Placeholder: overall \method{} architecture}\\[1mm]
    Replace with \texttt{figures/verifuse\_overview.pdf} (preferred) or
    \texttt{figures/verifuse\_overview.png}.\\[1mm]
    Suggested flow: detections $\rightarrow$ association $\rightarrow$ source-conditioned candidates
    $\rightarrow$ VLM action $\rightarrow$ deterministic compiler $\rightarrow$ one re-verification/fallback.}}
  }%
}
\caption{Overview of \method{}. Source-conditioned proposals form one mixed pool; the VLM selects, rejects, or returns one anchor--reason--token tuple, while deterministic compilation and re-verification bound geometric execution.}
\label{fig:pipeline}
\end{figure*}

Figure~\ref{fig:pipeline} summarizes the pipeline: common-frame proposals are associated, expanded into a bounded mixed pool, and passed to a frozen VLM. A \textsc{Refine} response identifies one anchor, reason code, and discrete token; a deterministic compiler instantiates one corrected box for re-verification or fallback.

\subsection{Association and Query Policy}
\label{sec:association}
For a vehicle proposal $v$ and an RSU proposal $r$, we minimize the gated cost
\begin{equation}
\begin{aligned}
\ell_{vr}={}&\lambda_{\rm I}
\left(1-\operatorname{IoU}_{\rm BEV}(b_v,b_r)\right)
+\lambda_{\rm p}\min\!\left(\frac{\|\vect p_v-\vect p_r\|}{R_{\rm p}},1\right)\\
&+\lambda_{\rm m}\min\!\left(
\frac{\|\widehat{\vect p}_v^t-\widehat{\vect p}_r^t\|}{R_{\rm m}},1\right),
\end{aligned}
\label{eq:association}
\end{equation}
using Hungarian matching. Incompatible classes or pairs outside the spatial gate have infinite cost. $\widehat{\vect p}^{\,t}$ is obtained by constant-velocity extrapolation from detector-track history no later than $t$. When history is unavailable, the motion term is removed and the remaining weights are renormalized. Additional agents are attached one source at a time to vehicle-anchored matches. A component containing two proposals from one source, or violating an all-pairs gate, is split by minimum-cost anchored matching. Thresholds and the compatibility map are frozen on the development set.

Vehicle-only clusters bypass VLM arbitration and retain their original vehicle proposal. Every cross-agent cluster and every infrastructure-only singleton enters the main arbitration pipeline. Apart from the mandatory vehicle-only pass-through, the full model has no additional consistency bypass. The fast path is evaluated only as a system ablation and applies its consistency bypass only after passing the same 300~ms timestamp gate as the full model. A failed timestamp check triggers deterministic fallback in both variants. \textsc{Reject} is admissible only when $g$ contains no vehicle proposal.

\subsection{Source-Conditioned Candidate Generation}
\label{sec:candidates}
Each candidate $h$ stores its geometry $\operatorname{box}(h)$, calibrated base score $\pi(h)$, class, provenance, and nonempty root set $\operatorname{roots}(h)\subseteq g$. A candidate is \emph{single-source} when $|\operatorname{roots}(h)|=1$; only then is $\operatorname{src}(h)$ defined.

For a perturbation $\vect\delta=(\vect\delta_{\rm p},\vect\delta_{\rm d},\delta_\psi)$, the generator uses the fixed box-update program
\begin{equation}
(\vect p,\vect d,\psi)\oplus\vect\delta=
\left(\vect p+\vect\delta_{\rm p},\;
\vect d\odot(\vect 1+\vect\delta_{\rm d}),\;
\operatorname{wrap}(\psi+\delta_\psi)\right).
\label{eq:box_update}
\end{equation}
Center offsets are expressed in the common fusion frame, dimension offsets are relative, and nonpositive dimensions are invalid. For original proposal $\nu=(a,i)$, the initial single-source hypotheses are
\begin{equation}
\begin{aligned}
\mathcal H_\nu^0={}&\{h_{\nu,0}\}\\[-1mm]
&\cup\{h_{\nu,\xi}:\vect\xi\in
\mathcal U_a^0\setminus\{\vect 0\}\}.
\end{aligned}
\label{eq:sourcehyp}
\end{equation}
Here $\operatorname{box}(h_{\nu,0})=b_\nu$ and $\operatorname{box}(h_{\nu,\xi})=b_\nu\oplus\Lambda_a\vect\xi$. A generated hypothesis is retained only when its box lies in $\mathcal N_\nu(b_\nu)$. The finite normalized lattice $\mathcal U_a^0$ is sparse, while diagonal $\Lambda_a$ converts each entry to source-specific center, relative-size, and yaw changes. Every $h_{\nu,\xi}$ has root set $\{\nu\}$, inherits class $c_\nu$, uses base score $\kappa_a(s_\nu)$, and records $(\nu,\xi)$ as provenance. The original is inserted explicitly. The lattice and scales are fitted on training and development residuals, then frozen without test-time learning or sampling.

For a matched vehicle--RSU pair, the full model also constructs deterministic cross-source boxes. A fixed finite set $\Omega\subset[0,1]$ supplies interpolation weights. For each $\omega\in\Omega$, the vehicle box has weight $\omega$ and the RSU box has weight $1-\omega$; centers and dimensions are linearly interpolated, and yaw follows the wrapped shortest arc. An additional confidence-weighted box uses
\begin{equation}
\omega_s=\frac{\kappa_V(s_v)}{\kappa_V(s_v)+\kappa_I(s_r)},
\label{eq:score_weight}
\end{equation}
with the zero-denominator case mapped to $1/2$. These boxes form $\mathcal H_g^{\rm cross}$, record both roots, and must remain in $\mathcal N(g)$. They may be selected but cannot anchor refinement because they have no unique source prior.

The final mixed pool is
\begin{equation}
\begin{aligned}
\mathcal H_g=\operatorname{Permute}_{\sigma(g)}\Big[&
\operatorname{Cap}^{\rm bal}_{K_{\max}}\big(
\operatorname{Dedup}(\\[-1mm]
&\bigcup_{\nu\in g}\mathcal H_\nu^0
\cup\mathcal H_g^{\rm cross})\big)\Big].
\end{aligned}
\label{eq:pool}
\end{equation}
All originals are reserved first. The remaining slots are filled round-robin from vehicle-derived, RSU-derived, and cross-source families in their frozen enumeration order, so the cap cannot silently eliminate one source. The deterministic permutation seed $\sigma(g)$ is derived from the sample and cluster identifiers and prevents order from encoding a preferred source.

The number of candidates is therefore deliberately variable. A singleton has one root, a matched cluster has two, cross-source boxes exist only for compatible matches, validity and deduplication may remove entries, and $K_{\max}$ truncates only large pools. These facts explain why different scene clusters and different generator ablations have different average candidate counts.

\subsection{Structured VLM Arbitration}
\label{sec:arbitration}
The frozen VLM receives $\mathcal H_g$ and timestamp-consistent evidence $E_g^t$: a metric BEV rendering of available point evidence, candidate overlays, and synchronized camera crops when available. It returns exactly one schema-validated record:
\begin{align}
\textsc{Select}&(j,\texttt{ADEQUATE},r),\nonumber\\
\textsc{Refine}&(j,q,\tau,r),\label{eq:contract}\\
\textsc{Reject}&(q,r),\nonumber
\end{align}
where $j$ is a candidate index, $q$ is an action-specific reason code, $\tau$ is a discrete correction-token identifier, and $r$ is a concise evidence summary used only for logging. Free-form text is never parsed, embedded, or executed.

The prompt first asks whether the evidence supports an object and then whether an adequate candidate exists. If the object is supported and an adequate candidate exists, the VLM returns \textsc{Select}; if the object is supported but every candidate is inadequate, it returns \textsc{Refine}; if the object is unsupported, \textsc{Reject} is available only to an infrastructure-only cluster. \texttt{ADEQUATE} is an inference-time visual criterion: the overlay has no visible center, yaw, or extent mismatch large enough to require an available correction. It is not an oracle IoU calculation; selection success is evaluated separately at the stated IoU threshold. For \textsc{Refine}, $j$ must index a single-source candidate and $\tau$ must belong to the source- and diagnosis-specific finite token set $\mathcal T_{\operatorname{src}(h_j)}(q)$.

The validator rejects missing indices, inadmissible actions, unknown reason codes, and incompatible or out-of-range tokens. Evidence sufficiency is checked before inference using deterministic timestamp, crop-bound, and sensor-support tests rather than a VLM self-reported confidence. If this gate fails, VLM inference is skipped and the cluster is handled by the fallback policy. We use the Qwen3-VL-8B-Instruct model~\cite{qwen3vl}, FP16 inference, and greedy decoding. Input resolutions and gates are listed in Table~\ref{tab:implementation}.

\subsection{Deterministic Token-to-Box Compiler}
\label{sec:compiler}
For \textsc{Select}, the output geometry is $\widetilde b_g=\operatorname{box}(h_j)$. For \textsc{Refine}, let $\nu_j$ be the unique root of $h_j$ and $a_j=\operatorname{src}(h_j)$. The VLM does not emit a numeric coordinate. Instead, the finite lookup table $\vect u_a(q,\tau)\in\R^7$ decodes token ID $\tau$ into a normalized correction vector. A binary diagonal mask $M_q$ enables only the components licensed by $q$, such as center, yaw, size, or transformed-source compensation. The compiler computes
\begin{equation}
\vect\delta_{j,q,\tau}=\Lambda_{a_j}M_q\vect u_{a_j}(q,\tau),
\qquad
b_{j,q,\tau}^\Gamma=\operatorname{box}(h_j)\oplus\vect\delta_{j,q,\tau}.
\label{eq:compiler}
\end{equation}
The executable program is a fixed lookup--mask--scale--update chain: one token deterministically instantiates one compiled box, followed by one accept-or-fallback verification.

Let $b_{\nu_j}^{\rm root}=b_{\nu_j}$ and define
\begin{equation}
\operatorname{res}(b,b^0)=
\left[(\vect p-\vect p^0)^\top,
(\vect d\oslash\vect d^0-\vect 1)^\top,
\operatorname{wrap}(\psi-\psi^0)\right]^\top.
\label{eq:box_residual}
\end{equation}
The compiled box is admissible only if
\begin{equation}
\chi_{a_j}(b_{j,q,\tau}^\Gamma,b_{\nu_j}^{\rm root})=
\left\|B_{a_j}^{-1}
\operatorname{res}(b_{j,q,\tau}^\Gamma,b_{\nu_j}^{\rm root})\right\|_\infty\le1
\label{eq:rootbound}
\end{equation}
and $V(b_{j,q,\tau}^\Gamma,E_g^t)=1$. Here $B_a$ is a positive diagonal scale matrix whose entries are the maximum source-specific center, relative-dimension, and wrapped-yaw deviations from the original proposal. The validity predicate $V$ checks the dataset range, positive dimensions, yaw convention, and frozen sensor-support rule. Accordingly, $\mathcal N_\nu(b_\nu)$ consists of $b_\nu$ and boxes satisfying the corresponding source-specific root bound and $V$; the initial lattice is restricted to this same neighborhood. Measuring Eq.~\eqref{eq:rootbound} from the original root rather than the selected anchor prevents initial and refinement perturbations from accumulating beyond the allowed neighborhood.

\subsection{Re-verification, Scoring, and Fallback}
\label{sec:output_policy}
After the schema, root-bound, and validity checks pass, the compiled \textsc{Refine} overlay is submitted once to the same frozen VLM through a separate binary prompt. This pass may return only \texttt{ACCEPT} or \texttt{FALLBACK}; it cannot request another correction. If accepted, $\widetilde b_g=b_{j,q,\tau}^\Gamma$. A schema failure, failed validity or bound test, or negative re-verification invokes the same deterministic fallback.

Let $\kappa_a(s)$ denote the per-source isotonic score calibration. A single-source candidate rooted at $\nu$ has base score $\pi_j=\pi(h_j)=\kappa_a(s_\nu)$; a cross-source candidate has $\pi_j=\sum_m w_m\kappa_{a_m}(s_{\nu_m})$, using its construction weights $w_m\ge0$ with $\sum_mw_m=1$. Selection returns $\widetilde s_g=\pi_j$, whereas an accepted refinement returns
\begin{equation}
\widetilde s_g=\pi_j\exp\!\left[
-\lambda_{\rm ref}\chi_{a_j}(b_{j,q,\tau}^\Gamma,b_{\nu_j}^{\rm root})\right].
\label{eq:refined_score}
\end{equation}
The class is inherited from the single-source root or from the compatible class of a cross-source candidate.

Fallback is unique across all failure locations. A vehicle-supported cluster returns its original vehicle proposal. For an infrastructure-only cluster with RSU root $\nu$, fallback returns the original proposal when $\kappa_I(s_\nu)\ge\theta_{\rm fb}$ and returns $\varnothing$ otherwise. A valid \textsc{Reject} directly returns $\varnothing$ and is allowed only for an infrastructure-only cluster.

\subsection{Execution and Communication Accounting}
\label{sec:execution}
The fusion node runs at the vehicle or an edge server. Agents transmit detection lists every frame. For a queried cluster, the RSU serializes a fixed-policy ROI evidence crop locally and sends it on demand; vehicle evidence is local to a vehicle-side fusion node. If inference runs at the edge, the structured action is returned as a control message. Offline evaluation reconstructs the same encoded messages and counts their serialized byte sizes as $B_{\rm total}=B_{\rm det}+B_{\rm evid}+B_{\rm ctrl}$. Trigger-rate-weighted evidence bytes are included, and centrally cached evidence is reported as local access rather than zero traffic.

\section{Experiments}
\label{sec:experiments}

We evaluate \method{} at both the detection and system levels. The end-to-end comparison measures cooperative detection while verifying that reliable vehicle proposals are preserved. Controlled ablations then separate the effects of candidate construction, VLM arbitration, and bounded refinement, followed by communication-aware latency and delayed-evidence tests. This ordering distinguishes whether an error originates in the available hypotheses, the arbitration decision, or the final correction. Tables~\ref{tab:main} and~\ref{tab:arbitration} retain the detailed method comparisons, whereas Figure~\ref{fig:overview} summarizes candidate coverage, refinement, and delay robustness. Code is available at \url{https://github.com/VeriFuse-Anonymous/VeriFuse}.

\subsection{Experimental Setup}
Experiments use DAIR-V2X-C~\cite{dairv2x}. Complete contiguous sequences separate detector fitting, development, and test data so that near-duplicate frames do not cross partitions. The development set fixes score calibration, association gates, perturbation supports, prompts, and all thresholds. A fixed, method-independent scene-content filter excludes the 56 frames with no annotated non-ego vehicle from the 600-frame held-out split. The resulting 544-frame test set is fixed before comparison and used identically by every primary method. Delay tests use a separate 2,000-frame sequence-disjoint stream excluded from fitting and configuration; 1,592 frames are valid at every offset.

For the primary detection study, we report 3D AP at IoU thresholds of 0.5 and 0.7, recall at 0.5 false positives per frame (R@0.5FP), and false positives per frame (FP/fr.) at each method's operating threshold frozen on the development set.
The delay study separately reports vehicle-side BEV AP$_{50}$. Together, these metrics distinguish loose detection coverage, strict localization, and false-positive control. To separate candidate availability from arbitration quality, we first test whether the initial pool contains a sufficiently accurate box. For positive clusters $\mathcal G_+$, let $b_g^*$ denote the matched ground-truth box. Initial-pool oracle coverage is
\begin{equation}
\mathrm{OC}_{\zeta}^{0}=\frac{1}{|\mathcal G_+|}
\sum_{g\in\mathcal G_+}\mathbf 1\!\left[
\max_{h\in\mathcal H_g}\operatorname{IoU}
(\operatorname{box}(h),b_g^*)\ge\zeta\right].
\label{eq:coverage}
\end{equation}
Let $\mathcal J_g^{\rm ss}$ index the single-source candidates and $\mathcal Q_{\rm ref}$ denote the refinement reason codes. Let $\mathcal A_g^{\rm valid}$ contain the triples $(j,q,\tau)$ for which $j\in\mathcal J_g^{\rm ss}$, $q\in\mathcal Q_{\rm ref}$, $\tau\in\mathcal T_{a_j}(q)$, and the compiled box $b_{j,q,\tau}^{\Gamma}$ satisfies the root-bound and validity tests. The one-step reachable geometries are
\begin{equation}
\begin{aligned}
\mathcal R_g={}&\{\operatorname{box}(h):h\in\mathcal H_g\}\\
&\cup\{b_{j,q,\tau}^{\Gamma}:
(j,q,\tau)\in\mathcal A_g^{\rm valid}\}.
\end{aligned}
\label{eq:reachable_set}
\end{equation}
Reachable coverage $\mathrm{OC}_{\zeta}^{\Gamma}$ replaces the maximum in Eq.~\eqref{eq:coverage} by $\max_{b\in\mathcal R_g}\operatorname{IoU}(b,b_g^*)$. This offline-only set upper-bounds one-step compiler reachability, while online inference instantiates only the VLM-selected token. Actual outputs are evaluated separately by selection success, refinement success, and IoU regret.

\begin{table}[t]
\centering
\caption{Fixed implementation settings.}
\label{tab:implementation}
\footnotesize
\setlength{\tabcolsep}{3pt}
\renewcommand{\arraystretch}{0.92}
\begin{tabularx}{\columnwidth}{@{}>{\raggedright\arraybackslash}p{0.37\columnwidth}>{\raggedright\arraybackslash}X@{}}
\toprule
Setting & Fixed value \\
\midrule
Data split & 1,200 / 200 / 600 paired frames; 544 test frames after fixed ego-only filtering \\
Detectors & Separate PointPillars~\cite{pointpillars}; best development 3D AP$_{50}$ checkpoint \\
Range / class & $[0,100]\!\times\![-40,40]\!\times\![-3,1]$ m; Car \\
Association & $(\lambda_{\rm I},\lambda_{\rm p},\lambda_{\rm m}){=}(0.45,0.35,0.20)$; $R_{\rm p}{=}3$ m, $R_{\rm m}{=}4$ m; IoU gate 0.10; 3 frames \\
Perturbation / bounds & Vehicle: 0.25 m / $3^\circ$ / 3\%; RSU: 0.75 m / $6^\circ$ / 6\%; signed half-step tokens; root-center bounds 0.75/1.50 m \\
Candidate cap & $K_{\max}{=}16$; originals-first, source-balanced truncation \\
VLM & Frozen Qwen3-VL-8B-Instruct, FP16, greedy, $T{=}0$ \\
Evidence & $896{\times}896$ BEV; $448{\times}448$ crops; 60 m ROI; JPEG quality 85 \\
Evidence gate / seed & Past-only warp, $|\Delta t|{\leq}300$ ms; $\geq8$ points or valid crop; CRC32 sample--cluster seed \\
Calibration / penalty & Per-source isotonic; $\lambda_{\rm ref}{=}0.35$ \\
Fallback threshold & $\theta_{\rm fb}{=}0.62$ \\
Platform & RTX 4090 24 GB; PyTorch 2.5.1; CUDA 12.4 \\
\bottomrule
\end{tabularx}
\end{table}

\subsection{Baselines and Main Results}
Table~\ref{tab:main} compares independent detectors and early-, intermediate-, and late-fusion baselines with two VLM variants on identical frames. The learned geometric selector is a supervised candidate ranker trained on the detector-fitting split using only calibrated detector scores and geometric/source features, including source identity, association cost, and relative center, size, and yaw. It receives no BEV or camera evidence, ranks the original clustered proposals, and uses a learned support score to reject unsupported infrastructure-only clusters. Its operating threshold is frozen on the development set. Table~\ref{tab:main} reports this end-to-end original-proposal configuration, whereas Table~\ref{tab:arbitration} supplies the same selector with the fixed full candidate pool to isolate arbitration quality. The independent detectors isolate the contribution of each source, early and intermediate fusion represent measurement- and feature-level cooperation, and the late-fusion rows test detector-independent result fusion. The \textsc{Select}/\textsc{Reject} variant isolates fixed-pool VLM arbitration, whereas the full model additionally activates bounded refinement. Both VLM variants share the frozen model, evidence, candidate ordering, and decoding settings, and use the same arbitration prompt template except for the permitted action set: the ablation disables \textsc{Refine}, whereas the full model enables it and uses the separate one-pass re-verification prompt described in Sec.~\ref{sec:output_policy}. The first two metrics use vehicle-side ground truth; the remaining metrics use cooperative ground truth. The comparison evaluates complete systems, including \method{}'s detection lists and on-demand ROI evidence.

\begin{table*}[!t]
\centering
\caption{Detection performance on the fixed 544-frame test set after ego-only scene filtering.}
\label{tab:main}
\footnotesize
\setlength{\tabcolsep}{3.0pt}
\begin{tabularx}{\textwidth}{@{}lLcccccc@{}}
\toprule
& & \multicolumn{2}{c}{Vehicle-side}
& \multicolumn{4}{c}{Cooperative} \\
\cmidrule(lr){3-4}\cmidrule(l){5-8}
Level & Method
& 3D AP$_{50}\uparrow$
& 3D AP$_{70}\uparrow$
& 3D AP$_{50}\uparrow$
& 3D AP$_{70}\uparrow$
& R@0.5FP$\uparrow$
& FP/fr.$\downarrow$ \\
\midrule
None & Vehicle detector
& 0.636 & 0.528 & 0.456 & 0.312 & 0.477 & \textbf{0.37} \\
None & RSU detector
& 0.421 & 0.283 & 0.401 & 0.242 & 0.292 & 2.18 \\
Early & Point-cloud concatenation
& 0.455 & 0.301 & 0.335 & 0.205 & 0.286 & 2.94 \\
\midrule
Intermediate & Where2comm~\cite{where2comm}
& 0.617 & 0.406 & 0.517 & 0.315 & 0.438 & 4.83 \\
Intermediate & CoAlign~\cite{coalign}
& 0.628 & 0.432 & 0.529 & 0.342 & 0.462 & 3.76 \\
Intermediate & HEAL~\cite{heal}
& 0.635 & 0.446 & \textbf{0.535} & 0.351 & 0.474 & 3.21 \\
\midrule
Late & NMS
& 0.620 & 0.515 & 0.476 & 0.327 & 0.468 & 0.90 \\
Late & Soft-NMS~\cite{softnms}
& 0.623 & 0.518 & 0.480 & 0.329 & 0.421 & 5.60 \\
Late & Weighted boxes fusion~\cite{wbf}
& 0.607 & 0.496 & 0.469 & 0.316 & 0.464 & 0.87 \\
Late & Learned geometric selector
& 0.632 & 0.532 & 0.483 & 0.337 & 0.484 & 0.96 \\
\rowcolor{vfPurpleFill}
Late + VLM & \method{} (\textsc{Select}/\textsc{Reject})
& 0.639 & 0.542 & 0.486 & 0.342 & 0.504 & 0.84 \\
\rowcolor{vfGreenFill}
Late + VLM & \method{} (full)
& \textbf{0.646}
& \textbf{0.557}
& 0.494
& \textbf{0.357}
& \textbf{0.520}
& 0.79 \\
\bottomrule
\end{tabularx}
\par\vspace{8pt}
\end{table*}

\begin{figure*}[!t]
\centering
\setlength{\abovecaptionskip}{0pt}
\includegraphics[width=\textwidth]{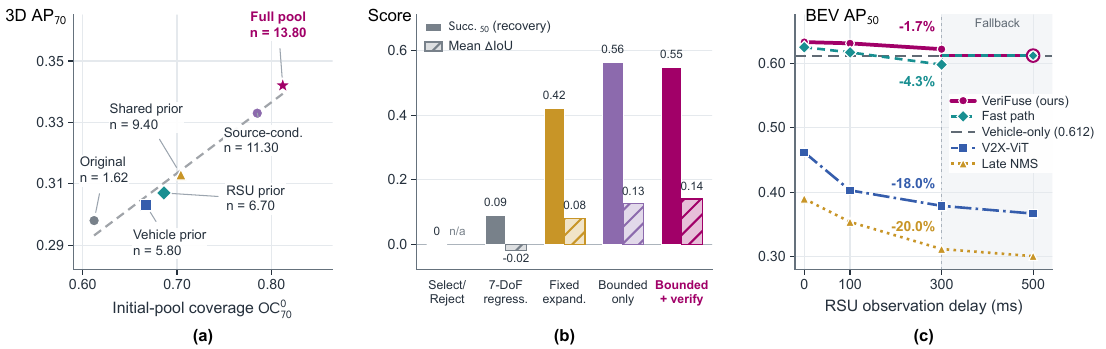}
\caption{Candidate coverage, refinement, and delay robustness of \method{}. (a)~Cooperative 3D AP$_{70}$ versus initial-pool oracle coverage $\mathrm{OC}_{70}^{0}$ with the same frozen VLM arbiter and refinement disabled; $n$ is the mean candidate count per cluster. (b)~Refinement on positive clusters with $\mathrm{OC}_{50}^{0}=0$: Succ.$_{50}$ measures recovery over the entire subset. (c)~Vehicle-side BEV AP$_{50}$ on the same 1,592 common valid frames at RSU delays of 0, 100, 300, and 500~ms. The gray dashed line marks the vehicle-only baseline (0.612). Percentage labels show relative AP decreases from 0 to 300~ms.}
\label{fig:overview}
\end{figure*}

Among the late-fusion approaches, the full \method{} model achieves the highest vehicle-side and cooperative 3D AP at both IoU thresholds, as well as the highest recall and the lowest false-positive rate. The full \textsc{Select}/\textsc{Refine}/\textsc{Reject} contract further improves upon fixed-pool \textsc{Select}/\textsc{Reject}: cooperative 3D AP$_{50}$/AP$_{70}$ increases from 0.486/0.342 to 0.494/0.357, recall increases from 0.504 to 0.520, and FP/fr. decreases from 0.84 to 0.79. The complete model also exceeds both independent detectors in cooperative 3D AP and recall. It improves vehicle-side 3D AP$_{50}$/AP$_{70}$ from 0.636/0.528 to 0.646/0.557 relative to the vehicle detector, even though the vehicle detector already performs strongly on this road segment.

\method{} outperforms Where2comm, CoAlign, and HEAL in cooperative 3D AP$_{70}$ while reducing FP/fr. to 0.79. Compared with HEAL, it achieves higher AP$_{70}$ and substantially fewer false positives, although its AP$_{50}$ is lower. This difference arises because feature-fusion methods aggregate dense cross-agent features and can recover additional objects, improving detection coverage under a looser IoU criterion. In contrast, \method{} operates within detector-supported geometric neighborhoods, focusing on correcting inaccurate boxes and suppressing unsupported proposals. This bounded optimization improves localization at the stricter IoU threshold while maintaining stronger false-positive control.

\subsection{Candidate Generation and Arbitration Ablations}
Figure~\ref{fig:overview}(a) varies candidate construction while fixing the VLM arbiter and disabling refinement. The full pool combines source-conditioned perturbations with cross-source hypotheses and achieves the highest initial-pool coverage $\mathrm{OC}^{0}_{70}$ and cooperative 3D AP$_{70}$. These results show that improved candidate coverage translates into more accurate localization.

Table~\ref{tab:arbitration} instead fixes the complete candidate pool and changes only the arbitration rule. Positive success is the fraction of positive queries selecting a box with IoU $\geq0.5$, and IoU regret is best-minus-selected IoU. Refinement is disabled in this comparison. Because every row receives the same hypotheses, the comparison isolates decision quality from candidate availability. Besides, relative to the learned selector, the VLM improves positive success, reduces regret, and raises 3D AP$_{70}$.

\begin{table}[!t]
\centering
\caption{Arbitration ablation on the fixed full candidate pool.}
\label{tab:arbitration}
\footnotesize
\setlength{\tabcolsep}{2.5pt}
\begin{tabularx}{\columnwidth}{@{}Lcccc@{}}
\toprule
Arbitration rule
& Pos.\ succ.$\uparrow$
& IoU regret$\downarrow$
& 3D AP$_{50}\uparrow$
& 3D AP$_{70}\uparrow$ \\
\midrule
Random selection
& 0.312 & 0.214 & 0.401 & 0.236 \\
Maximum confidence
& 0.598 & 0.126 & 0.452 & 0.291 \\
Minimum geom. cost
& 0.641 & 0.108 & 0.459 & 0.304 \\
Learned geom. selector
& 0.735 & 0.071 & 0.476 & 0.323 \\
\rowcolor{vfGreenFill}
VLM arbiter (ours)
& \textbf{0.836}
& \textbf{0.041}
& \textbf{0.486}
& \textbf{0.342} \\
\bottomrule
\end{tabularx}
\end{table}

\subsection{Refinement Ablation}
Figure~\ref{fig:overview}(b) and Table~\ref{tab:refinement} evaluate the hard subset of positive clusters whose initial candidates all have IoU below 0.5, so OC$_{50}^{0}=0$ by construction. The figure compares recovery and geometric improvement; the table retains precise values and the additional reachable-coverage and fallback metrics. The bounded variants attempt refinement on 84.7\% of these clusters. OC$_{50}^{\Gamma}$ is the offline upper bound for each finite correction set, Succ.$_{50}$ measures the actual output over the entire subset, $\Delta$IoU averages non-fallback outputs, and FB is fallback rate. Direct 7-DoF regression has no finite compiler support and therefore no reachable-coverage value.

\begin{table}[t]
\centering
\caption{Refinement ablation on the initially uncovered hard subset.}
\label{tab:refinement}
\footnotesize
\setlength{\tabcolsep}{4.0pt}
\begin{tabularx}{\columnwidth}{@{}Lcccc@{}}
\toprule
Method & OC$_{50}^{\Gamma}\uparrow$ & Succ.$_{50}\uparrow$ & $\Delta$IoU$\uparrow$ & FB(\%)$\downarrow$ \\
\midrule
\textsc{Select}/\textsc{Reject} & -- & 0.000 & -- & -- \\
VLM 7-DoF regression & -- & 0.091 & $-0.018$ & 23.4 \\
Fixed expansion & 0.691 & 0.421 & 0.081 & 0.0 \\
Bounded token w/o verify & 0.824 & 0.564 & 0.126 & 2.2 \\
\rowcolor{vfGreenFill}
Bounded token + verify & 0.824 & 0.548 & 0.141 & 7.4 \\
\bottomrule
\end{tabularx}
\end{table}

On initially uncovered positive clusters, bounded refinement with verification recovers 54.8\%, compared with only 9.1\% for direct 7-DoF VLM regression---approximately six times the recovery rate. Among non-fallback outputs, bounded refinement increases IoU by 0.141 on average, whereas direct regression decreases it by 0.018. This comparison demonstrates the central advantage of \method{}: the VLM is more effective at selecting a constrained geometric action than at generating box coordinates directly. Both bounded-token variants also outperform fixed expansion, which recovers 42.1\% of these clusters.

The two bounded-token variants share the same reason codes, token lookup, compiler, and fallback policy; only the final row enables re-verification. This additional check never modifies the compiled box. It filters questionable corrections, reducing recovery from 0.564 to 0.548 while increasing the mean IoU gain among retained outputs from 0.126 to 0.141 and raising fallback from 2.2\% to 7.4\%. Bounded action selection provides the main recovery gain, while re-verification trades some recovery for greater mean geometric improvement among accepted outputs.

\subsection{Runtime and Delay Robustness}

End-to-end latency accounts for detection, evidence rendering, VLM inference, response parsing, deterministic geometry, and communication. Under a 10~Mbps link, \method{} transmits 192.904~KB/frame and incurs a mean latency of 1,115~ms, compared with 16,794~KB/frame and 14,050~ms for V2X-ViT. These correspond to reductions of 98.9\% in payload and 92.1\% in communication-aware mean latency. Compact detection lists and on-demand ROI evidence concentrate communication on the information needed for arbitration, substantially reducing the cost of cooperation under the tested bandwidth constraint.

Figure~\ref{fig:overview}(c) evaluates delay robustness on the same 1,592 common valid frames, with vehicle observations held fixed. At 300~ms, \method{} retains 0.622 BEV AP$_{50}$, above the vehicle-only baseline of 0.612, with a 1.7\% relative decrease from zero delay. The fast path falls below vehicle-only to 0.598, with a 4.3\% decrease; V2X-ViT and late NMS decrease by 18.0\% and 20.0\%, respectively. Beyond the shared 300~ms timestamp gate, both variants invoke deterministic fallback and match vehicle-only AP at 500~ms. The fast path's recovery from 300 to 500~ms reflects this switch to fallback. Full arbitration preserves cooperation gains at the tested delays within the evidence window, while the shared fallback restores baseline-level AP at 500~ms.

\subsection{Qualitative Results}
\begin{figure*}[!t]
\centering
\IfFileExists{figures/verifuse_qualitative.pdf}{%
  \includegraphics[width=\textwidth]{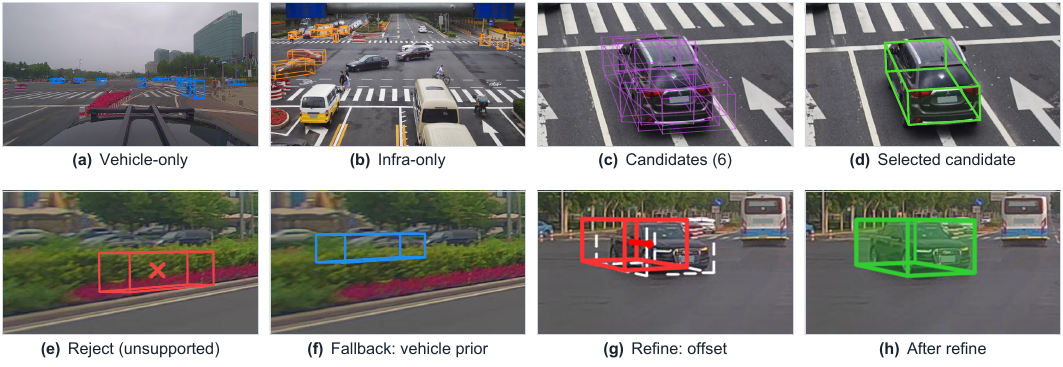}%
}{%
  \IfFileExists{figures/verifuse_qualitative.png}{%
    \includegraphics[width=\textwidth]{figures/verifuse_qualitative.png}%
  }{%
    \fbox{\parbox[c][0.19\textheight][c]{0.96\textwidth}{\centering
    \textbf{Placeholder: qualitative decision-path figure}\\[1mm]
    Replace with \texttt{figures/verifuse\_qualitative.pdf} (preferred) or
    \texttt{figures/verifuse\_qualitative.png}.\\[1mm]
    Recommended panels: vehicle-only, infrastructure-only, candidate pool,
    \textsc{Select}, \textsc{Reject} of a false positive, fallback, and
    refine before/after, each pairing the vehicle-view image with its BEV
    detections.}}
  }%
}
\caption{Illustration of \method{} on DAIR-V2X-C camera images. Box overlays illustrate the decision paths: (a)~vehicle-only and (b)~infrastructure-only detections; (c)~the mixed candidate pool; (d)~\textsc{Select} of an adequate candidate; (e,f) an inadmissible \textsc{Reject} response for a vehicle-supported cluster, followed by fallback to its original vehicle proposal; (g)~a \textsc{Refine} diagnosis of a center offset; and (h)~the refined box after deterministic compilation.}
\label{fig:qualitative}
\end{figure*}

Figure~\ref{fig:qualitative} illustrates the decision paths of \method{} on DAIR-V2X-C camera images. Panels (a,b) show vehicle-side and infrastructure-side detections. Panels (c,d) show the mixed candidate pool and the candidate chosen by \textsc{Select}. Panels (e,f) illustrate fallback following an inadmissible \textsc{Reject} response. Because the cluster contains a vehicle proposal, the validator discards the response in (e) and returns the original vehicle detection in (f). Panels (g,h) illustrate the \textsc{Refine} process, where the center-offset error in (g) is corrected by the deterministic compiler to produce the refined box in (h).

\section{Discussion and Limitations}
\label{sec:discussion}

The refinement ablation establishes bounded action selection as a more effective role for the VLM than direct coordinate regression in the evaluated cooperative-perception setting. Constrained decisions recover accurate boxes from initially inadequate candidate pools, whereas direct regression achieves low recovery and a negative mean IoU change among non-fallback outputs. Specialized detectors provide metric anchors, the VLM identifies an appropriate correction, and the deterministic compiler executes that correction under fixed geometric constraints. This division of labor makes semantic reasoning useful for geometric perception while retaining explicit control over the final box.

These constraints also define the method's limitations. Because every output remains rooted in an existing detection, \method{} cannot recover objects missed by all agents or correct global calibration and association errors. Moreover, performance depends on detector quality, sensor configuration, and the predefined token space. Additionally, current VLM latency remains unsuitable for hard real-time control. Future work should investigate lightweight query policies, temporal uncertainty reasoning, and adaptive but verifiable action spaces. The central challenge is to expand the semantic contribution of foundation models without surrendering the reliability of metric execution.

\section{Conclusion}
We presented \method{}, a new paradigm for integrating VLMs into cooperative perception as bounded semantic optimizers. Specialized detectors generate metric hypotheses, a frozen VLM selects structured actions, and a deterministic compiler executes bounded geometric updates. The experimental results show that bounded token refinement recovers 54.8\% of initially uncovered positive clusters, compared with 9.1\% for direct 7-DoF VLM regression. Among non-fallback outputs, bounded refinement improves mean IoU, whereas direct regression worsens it. These results demonstrate the value of assigning VLMs constrained geometric decisions while retaining deterministic coordinate execution. The full system also improves strict localization and false-positive control. Under a 10~Mbps link, compact detection lists and on-demand ROI evidence reduce communication payload by 98.9\% and mean end-to-end latency by 92.1\% relative to V2X-ViT. This bounded decision--execution interface provides a foundation for extending semantic reasoning to future vision--language--action models in cooperative systems.

\flushend
\bibliographystyle{IEEEtran}
\bibliography{VLM}

@inproceedings{dairv2x,
  author    = {Yu, Haibao and Luo, Yizhen and Shu, Mao and Huo, Yiyi and Yang, Zebang and Shi, Yifeng and Guo, Zhenglong and Li, Hanyu and Hu, Xing and Yuan, Jirui and Nie, Zaiqing},
  title     = {{DAIR-V2X}: A Large-Scale Dataset for Vehicle-Infrastructure Cooperative {3D} Object Detection},
  booktitle = {Proc. IEEE/CVF Conf. Comput. Vis. Pattern Recognit. (CVPR)},
  pages     = {21361--21370},
  year      = {2022}
}

@inproceedings{v2vnet,
  author    = {Wang, Tsun-Hsuan and Manivasagam, Sivabalan and Liang, Ming and Yang, Bin and Zeng, Wenyuan and Urtasun, Raquel},
  title     = {{V2VNet}: Vehicle-to-Vehicle Communication for Joint Perception and Prediction},
  booktitle = {Proc. Eur. Conf. Comput. Vis. (ECCV)},
  pages     = {605--621},
  year      = {2020},
  doi       = {10.1007/978-3-030-58536-5_36}
}

@inproceedings{disconet,
  author    = {Li, Yiming and Ren, Shunli and Wu, Pengxiang and Chen, Siheng and Feng, Chen and Zhang, Wenjun},
  title     = {Learning Distilled Collaboration Graph for Multi-Agent Perception},
  booktitle = {Adv. Neural Inf. Process. Syst. (NeurIPS)},
  volume    = {34},
  pages     = {29541--29552},
  year      = {2021}
}

@inproceedings{v2xvit,
  author    = {Xu, Runsheng and Xiang, Hao and Tu, Zhengzhong and Xia, Xin and Yang, Ming-Hsuan and Ma, Jiaqi},
  title     = {{V2X-ViT}: Vehicle-to-Everything Cooperative Perception with Vision Transformer},
  booktitle = {Proc. Eur. Conf. Comput. Vis. (ECCV)},
  pages     = {107--124},
  year      = {2022},
  doi       = {10.1007/978-3-031-19842-7_7}
}

@inproceedings{where2comm,
  author    = {Hu, Yue and Fang, Shaoheng and Lei, Zixing and Zhong, Yiqi and Chen, Siheng},
  title     = {{Where2comm}: Communication-Efficient Collaborative Perception via Spatial Confidence Maps},
  booktitle = {Adv. Neural Inf. Process. Syst. (NeurIPS)},
  volume    = {35},
  pages     = {4874--4886},
  year      = {2022}
}

@inproceedings{cobevt,
  author    = {Xu, Runsheng and Tu, Zhengzhong and Xiang, Hao and Shao, Wei and Zhou, Bolei and Ma, Jiaqi},
  title     = {{CoBEVT}: Cooperative Bird's Eye View Semantic Segmentation with Sparse Transformers},
  booktitle = {Proc. Conf. Robot Learn. (CoRL)},
  series    = {Proceedings of Machine Learning Research},
  volume    = {205},
  pages     = {989--1000},
  publisher = {PMLR},
  year      = {2023}
}

@inproceedings{coalign,
  author    = {Lu, Yifan and Li, Quanhao and Liu, Baoan and Dianati, Mehrdad and Feng, Chen and Chen, Siheng and Wang, Yanfeng},
  title     = {Robust Collaborative {3D} Object Detection in Presence of Pose Errors},
  booktitle = {Proc. IEEE Int. Conf. Robot. Autom. (ICRA)},
  pages     = {4812--4818},
  year      = {2023},
  doi       = {10.1109/ICRA48891.2023.10160546}
}

@inproceedings{heal,
  author    = {Lu, Yifan and Hu, Yue and Zhong, Yiqi and Wang, Dequan and Wang, Yanfeng and Chen, Siheng},
  title     = {An Extensible Framework for Open Heterogeneous Collaborative Perception},
  booktitle = {Proc. Int. Conf. Learn. Represent. (ICLR)},
  year      = {2024}
}

@inproceedings{stamp,
  author    = {Gao, Xiangbo and Xu, Runsheng and Li, Jiachen and Wang, Ziran and Fan, Zhiwen and Tu, Zhengzhong},
  title     = {{STAMP}: Scalable Task- and Model-Agnostic Collaborative Perception},
  booktitle = {Proc. Int. Conf. Learn. Represent. (ICLR)},
  year      = {2025}
}

@inproceedings{softnms,
  author    = {Bodla, Navaneeth and Singh, Bharat and Chellappa, Rama and Davis, Larry S.},
  title     = {Soft-{NMS}---Improving Object Detection with One Line of Code},
  booktitle = {Proc. IEEE Int. Conf. Comput. Vis. (ICCV)},
  pages     = {5561--5569},
  year      = {2017},
  doi       = {10.1109/ICCV.2017.593}
}

@article{wbf,
  author  = {Solovyev, Roman and Wang, Weimin and Gabruseva, Tatiana},
  title   = {Weighted Boxes Fusion: Ensembling Boxes from Different Object Detection Models},
  journal = {Image Vis. Comput.},
  volume  = {107},
  pages   = {104117},
  year    = {2021},
  doi     = {10.1016/j.imavis.2021.104117}
}

@inproceedings{drivelm,
  author    = {Sima, Chonghao and Renz, Katrin and Chitta, Kashyap and Chen, Li and Zhang, Hanxue and Xie, Chengen and Bei{\ss}wenger, Jens and Luo, Ping and Geiger, Andreas and Li, Hongyang},
  title     = {{DriveLM}: Driving with Graph Visual Question Answering},
  booktitle = {Computer Vision -- ECCV 2024},
  series    = {Lecture Notes in Computer Science},
  volume    = {15110},
  pages     = {256--274},
  publisher = {Springer Nature Switzerland},
  year      = {2025},
  doi       = {10.1007/978-3-031-72943-0_15}
}

@inproceedings{drivevlm,
  author    = {Tian, Xiaoyu and Gu, Junru and Li, Bailin and Liu, Yicheng and Wang, Yang and Zhao, Zhiyong and Zhan, Kun and Jia, Peng and Lang, XianPeng and Zhao, Hang},
  title     = {{DriveVLM}: The Convergence of Autonomous Driving and Large Vision-Language Models},
  booktitle = {Proc. Conf. Robot Learn. (CoRL)},
  series    = {Proceedings of Machine Learning Research},
  volume    = {270},
  pages     = {4698--4726},
  publisher = {PMLR},
  year      = {2025}
}

@inproceedings{omnidrive,
  author    = {Wang, Shihao and Yu, Zhiding and Jiang, Xiaohui and Lan, Shiyi and Shi, Min and Chang, Nadine and Kautz, Jan and Li, Ying and Alvarez, Jose M.},
  title     = {{OmniDrive}: A Holistic Vision-Language Dataset for Autonomous Driving with Counterfactual Reasoning},
  booktitle = {Proc. IEEE/CVF Conf. Comput. Vis. Pattern Recognit. (CVPR)},
  pages     = {22442--22452},
  year      = {2025}
}

@misc{lin2026proxyselect,
  author        = {Lin, Hongyi and Liu, Yang and Zhao, Jinhua and Qu, Xiaobo},
  title         = {Rethinking Foundation Model Collaboration: Enhancing Specialized Models through Proxy Task Reasoning},
  year          = {2026},
  eprint        = {2606.31157},
  archivePrefix = {arXiv},
  primaryClass  = {cs.CV},
  doi           = {10.48550/arXiv.2606.31157}
}

@misc{zhou2026sotif,
  author        = {Zhou, Ji and Ding, Yilin and Zhao, Yongqi and Xu, Jiachen and Eichberger, Arno},
  title         = {A Comparative Evaluation of Large Vision-Language Models for {2D} Object Detection under {SOTIF} Conditions},
  year          = {2026},
  eprint        = {2601.22830},
  archivePrefix = {arXiv},
  primaryClass  = {cs.CV},
  doi           = {10.48550/arXiv.2601.22830}
}

@article{paetzold2025vlmgist,
  author  = {P{\"a}tzold, Bastian and Nogga, Jan and Behnke, Sven},
  title   = {Leveraging Vision-Language Models for Open-Vocabulary Instance Segmentation and Tracking},
  journal = {IEEE Robot. Autom. Lett.},
  volume  = {10},
  number  = {11},
  pages   = {11578--11585},
  month   = nov,
  year    = {2025},
  doi     = {10.1109/LRA.2025.3606363}
}

@inproceedings{pointpillars,
  author    = {Lang, Alex H. and Vora, Sourabh and Caesar, Holger and Zhou, Lubing and Yang, Jiong and Beijbom, Oscar},
  title     = {{PointPillars}: Fast Encoders for Object Detection from Point Clouds},
  booktitle = {Proc. IEEE/CVF Conf. Comput. Vis. Pattern Recognit. (CVPR)},
  pages     = {12697--12705},
  year      = {2019}
}

@article{qwen3vl,
  author        = {Bai, Shuai and others},
  title         = {{Qwen3-VL} Technical Report},
  journal       = {arXiv preprint arXiv:2511.21631},
  year          = {2025},
  eprint        = {2511.21631},
  archivePrefix = {arXiv},
  primaryClass  = {cs.CV}
}

@article{lin2025big,
  title={Big data-driven advancements and future directions in vehicle perception technologies: From autonomous driving to modular buses},
  author={Lin, Hongyi and Liu, Yang and Wang, Liang and Qu, Xiaobo},
  journal={IEEE Transactions on Big Data},
  volume={11},
  number={3},
  pages={1568--1587},
  year={2025},
  publisher={IEEE}
}

\end{document}